\documentclass{article}

\usepackage{microtype}
\usepackage{graphicx}
\usepackage{subfigure}
\usepackage{booktabs} 

\usepackage{hyperref}

\usepackage{amsmath,amsfonts,bm}

\def\eqref#1{equation~\ref{#1}}

\def\1{\bm{1}}

\DeclareMathAlphabet{\mathsfit}{\encodingdefault}{\sfdefault}{m}{sl}
\SetMathAlphabet{\mathsfit}{bold}{\encodingdefault}{\sfdefault}{bx}{n}

\usepackage{hyperref}
\usepackage{url}
\usepackage{tabularx}
\usepackage{wrapfig}
\usepackage[export]{adjustbox}
\usepackage{csquotes}
\usepackage{multirow}

\usepackage{tikz}
\usepackage{tikzscale}
\usepackage{tikz-dependency}

\usepackage{amsmath,amsfonts,amssymb}

\newcommand{\secc}[1]{Section~\ref{sec:#1}}
\newcommand{\fig}[1]{Figure~\ref{fig:#1}}
\newcommand{\tab}[1]{Table~\ref{tab:#1}}
\newcommand{\eq}[1]{Eq.~\ref{eq:#1}}

\renewcommand{\vec}{\mathbf}

\usepackage[accepted]{icml2021}

\icmltitlerunning{Learning to Forget by Expiring}

\begin{document}

\twocolumn[
\icmltitle{Not All Memories are Created Equal: \\ Learning to Forget by Expiring}




\begin{icmlauthorlist}
\icmlauthor{Sainbayar Sukhbaatar}{fb}
\icmlauthor{Da Ju}{fb}
\icmlauthor{Spencer Poff}{fb}
\icmlauthor{Stephen Roller}{fb}
\icmlauthor{Arthur Szlam}{fb}
\icmlauthor{Jason Weston}{fb}
\icmlauthor{Angela Fan}{fb,loria}
\end{icmlauthorlist}

\icmlaffiliation{fb}{Facebook AI Research}
\icmlaffiliation{loria}{LORIA}

\icmlcorrespondingauthor{Sainbayar Sukhbaatar}{sainbar@fb.com}

\icmlkeywords{Memory, Expiration, Transformers}

\vskip 0.3in
]



\printAffiliationsAndNotice{} 

\begin{abstract}
Attention mechanisms have shown promising results in sequence modeling tasks that require long-term memory. Recent work investigated mechanisms to reduce the computational cost of preserving and storing memories \citep{rae2020compressive}. However, not all content in the past is equally important to remember. We propose \textit{Expire-Span}, a method that learns to retain the most important information and \textit{expire} the irrelevant information. This forgetting of memories enables Transformers to scale to attend over tens of thousands of previous timesteps efficiently, as not all states from previous timesteps are preserved. We demonstrate that Expire-Span can help models identify and retain critical information and show it can achieve strong performance on reinforcement learning tasks specifically designed to challenge this functionality. Next, we show that Expire-Span can scale to memories that are tens of thousands in size, setting a new state of the art on incredibly long context tasks such as character-level language modeling and a frame-by-frame moving objects task. Finally, we analyze the efficiency of Expire-Span compared to existing approaches and demonstrate that it trains faster and uses less memory.
\end{abstract}

\newcommand{\fix}{\marginpar{FIX}}
\newcommand{\new}{\marginpar{NEW}}

\section{Introduction}

Transformer architectures~\citep{vaswani2017attention} have demonstrated strong performance across a variety of tasks~\citep{devlin2019bert,roller2020recipes,brown2020language}, including those that require learning long term relationships~\citep{zhang2018improving,fan2019using,izacard2020leveraging}. 
Recent work has focused on scaling attention mechanisms efficiently to longer memory sizes, enabling large improvements on long context tasks~\citep{dai2019transformer,sukhbaatar2019adaptive}. However, a critical component of human memory is not just the ability to remember, but also \textit{forgetting} irrelevant information to focus on the salient, relevant bits. Most studies of long-term memory in humans indicate that not everything is remembered~\citep{Murre2015ReplicationAA,bahrick2008fifty} --- instead, only vivid, remarkable memories are retained from the far past~\citep{wixted2004psychology}. 

Standard Transformer architectures lack the ability to search over extremely large memories, as the self-attention mechanism is computationally intensive and the storage cost of preserving the large memory grows quickly. Recent work~\citep{child2019generating,rae2020compressive} has proposed learning how to extend to greater context through sparse mechanisms or through compression, to more compactly represent the past. However, there exists a fundamental problem with large memories beyond strict computational concerns: as the amount of information stored increases, deciding which  information is relevant becomes more challenging.  Other work~\citep{lample2019large} approaches this by considering how to efficiently search large memories.  We focus on an efficient way to learn what to forget, thereby reducing the computational burden of the model and easing the challenges of the search problem.

We propose \textsc{Expire-Span}, a straightforward extension to attention mechanisms that learns when to \textit{expire} unneeded memories. By expiring memories that are no longer useful, \textsc{Expire-Span} enables scaling to tens of thousands of timesteps into the past. This learnable mechanism allows the model to adjust the span size as needed, selecting which information is critical to retain and forgetting the rest. 
More concretely, we augment the self-attention with a simple predictor that outputs an expiration value for each hidden state that determines how long a memory should be retained and accessible to the model.
After the \textsc{Expire-Span} runs out, the memory will be forgotten, but in a gradually differentiable way to retain end-to-end training with backpropagation.
This process is done independently for each layer, allowing different layers to specialize at different time-scales.
As \textsc{Expire-Span} can flexibly adjust its span based on context, it is more efficient in terms of memory and training time compared to existing long memory approaches. 

We demonstrate that \textsc{Expire-Span} can distinguish between critical and irrelevant information on several illustrative tasks in natural language processing and reinforcement learning that are specifically designed to test this ability. We then show we can achieve state-of-the-art results on long-context language modeling benchmarks, and \textsc{Expire-Span} can scale to memories in the tens of thousands on a frame-by-frame colliding objects task --- by expiring irrelevant information, capacity is freed to have even larger memory. Then, we compare the efficiency of our method to competitive baselines and show \textsc{Expire-Span} is faster and has a smaller memory footprint. Finally, we analyze the information retained and expired by \textsc{Expire-Span} models, to understand the importance of long context memory.

\section{Related Work}
Memory is crucial for many tasks and has been studied in recurrent networks~\citep{Elman1990FindingSI,hochreiter1997long,mikolov2010recurrent} for a long time.
The development of memory augmented networks~\citep{graves2014neural,sukhbaatar2015end} made it possible to store large quantities of information and selectively access them using attention~\citep{bahdanau2014neural}. 
The Transformer~\citep{vaswani2017attention} took full advantage of this approach.
Processing long sequences with Transformers is an active area with applications in language understanding~\citep{brown2020language}, reinforcement learning~\citep{Parisotto2019StabilizingTF}, video processing~\citep{wu2019long}, and protein folding~\citep{Rives2019BiologicalSA,choromanski2020masked}. 
However, extending the memory span is computationally expensive due to the quadratic time and space complexity of self-attention.
Other work focuses on benchmarking long memories~\citep{tay2020long}, but focuses on encoder-only tasks, whereas we focus on decoder-only Transformers.

Various work has focused on reducing this complexity and increasing memory capacity~\citep{schlag2021linear}. Dynamic attention spans, such as Adaptive-Span~\citep{sukhbaatar2019adaptive} and Adaptively Sparse Transformer~\citep{correia2019adaptively}, focus on learning which attention heads can have shorter spans, but can only extend to spans of a few thousand. Other work sparsifies attention by computing fewer tokens~\citep{fan2019strategies}, often by using fixed attention masks~\citep{child2019generating} or sliding windows and dilation ~\citep{beltagy2020longformer}. The BP Transformer~\citep{ye2019bp} structures tokens as a tree, so some tokens have coarse attention. These works focus on learning what to attend to, but searching larger and larger memories is very difficult. In contrast, we focus on learning to expire what is irrelevant. 
Compressive Transformer \citep{rae2020compressive} reduces the number of memories by replacing every few memories with a single compressed one. A disadvantage of this is that all memories have the same compression ratio, so relevant memories are equally compressed.

Another line of work investigates linear-time attention mechanisms. \citet{wu2018pay} replace self-attention with convolutions that run in linear time, but the scalability to long context tasks remains limited. \citet{wang2020linformer} propose linear time attention by decomposing attention into multiple smaller attentions, that recombine to form a low-rank factorization of the original attention.  
\citet{katharopoulos2020transformers} propose linear attention by expressing self-attention as instead a linear dot-product of feature maps.
\citet{peng2021random} propose Random Feature Attention, used to approximate the softmax.
Those methods, however, focus on making attention more efficient without reducing the number of memories. Further, as our goal is to reduce the number of memories that feed to self-attention by learning to expire, \textsc{Expire-Span} can be easily combined with these efficiency improvements.
For a review of further recent Transformer variants, see 
\citet{tay2020efficient}.

\section{Background}

\label{sec:back}
\newcommand{\inputTikZ}[2]{%
     \scalebox{#1}{\input{#2}}  
}

\begin{figure}[t]
        \centering
        \vspace{2mm}
     \scalebox{0.94}{\pgfdeclarelayer{background}
\pgfsetlayers{background,main}

\begin{tikzpicture}[
>=stealth',
]
\coordinate (O) at (0,0);
\draw[->,thick] (0,-0.5) -- (6.5,-0.5) coordinate[label = {below: time}] (xmax);

\node at (0.6,2.7) {Expire-span};
\begin{scope}[on background layer]
    \filldraw[fill=gray!10!white] (-0.35,0) rectangle (5.5,3);
\end{scope}

\foreach \i/\l/\c in {
    1/4.0/green, 
    2/1.0/gray, 
    3/2.0/green, 
    4/1.2/green, 
    5/0.25/gray} {
    \begin{scope}[shift={(\i*0.5,\i*0.5)}]
        \draw (0,0) node[label = {left: $e_\i$}] {};
        \fill[\c!40!gray] (0,-0.1) rectangle (\l,0.1);
        \shade[left color=\c!40!gray,right color=gray!10!white] (\l-0.01,-0.1) rectangle (\l+0.5,0.1);
        \node[] (barA\i) at (0,0) {};
        \node[] (barB\i) at (\l,0) {};
        \node[] (barC\i) at (\l+0.5,0) {};
    \end{scope}                   
    \filldraw[fill=white] (\i*0.5,-0.5) circle (0.1) node[label = {below: $\vec{h}_\i$}] (hi) {};
    \draw[<-,thin] (\i*0.5,0) -- (hi);
    \begin{scope}[on background layer]
        \draw[dotted,thin] (barA\i) -- (\i*0.5,0);
    \end{scope}
}

        
\node[rectangle,thick,draw,inner sep=2mm] (attn) at (7,1.5) {Attention};
\draw [->] (barC1.east) to [out=0,in=180] (attn.west);
\draw [->] (barC3.east) to [out=0,in=180] (attn.west);
\draw [->] (barC4.east) to [out=0,in=180] (attn.west);

\node[node distance=0.7] (qt) [below=of attn] {$\vec{q}_t$};
\draw [->] (qt) --  node[auto,swap] {query} (attn) ;
\node[node distance=0.7] (ot) [above=of attn] {$\vec{o}_t$};
\draw [->] (attn) -- node[auto,swap] {output} (ot);

\filldraw[fill=white] (3.5,-0.5) circle (0.1)  node[label = {below: $\vec{h}_t$}] (ht) {};
\draw[dashed,thick] (ht) -- (3.5,3);
\draw [->] (ht.east) to [out=20,in=-150] (qt.west);



\end{tikzpicture} }  

        \vspace{-3mm}
        \caption{\textbf{Expire-Span}. For every memory $\vec{h}_i$, we compute an \textsc{Expire-Span} $e_i$ that determines how long it should stay in memory. Here, memories $\vec{h}_2$, $\vec{h}_5$ are already expired at time $t$, so the query $\vec{q}_t$ can only access $\{\vec{h}_1, \vec{h}_3, \vec{h}_4\}$ in self-attention.}
        \label{fig:model}
\end{figure}
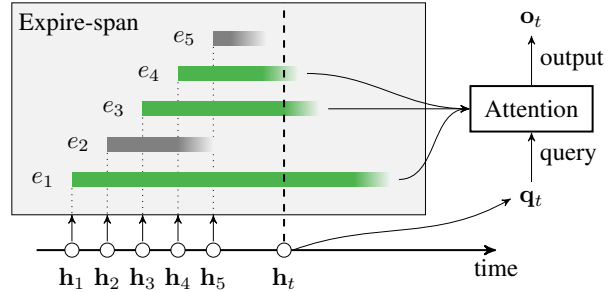

Transformer architectures have been widely used as decoder-only auto-regressive models for sequential tasks. A Transformer decoder is made of a stack of identical layers, composed of a multi-head self-attention sublayer followed by a feedforward sublayer. The output of each timestep is the hidden state $\vec{h}_t^l \in \mathbb{R}^d $ at layer $l$, which is then projected to key $\vec{k}$, value $\vec{v}$, and query $\vec{q}$ vectors:
\begin{equation}
\vec{q}_t^l = W_q^l \vec{h}_t^l, \quad \vec{k}_t^l = W_k^l \vec{h}_t^l, \quad \vec{v}_t^l = W_v^l \vec{h}_t^l .
\end{equation}
Going forward, we focus on a single layer and omit the layer index $l$ for brevity.
Information from previous timesteps is accessed through attention $a_{ti}$ to create output $\vec{o}_t$: 
\begin{equation}
a_{ti} = \underset{i \in C_t}{ \text{Softmax}} \left(\vec{q}_t^\top \vec{k}_i\right), \quad \vec{o}_t = W_o \sum_{i \in C_t} a_{t,i} \vec{v}_i.
\end{equation}
The set $C_t$ indicates which memories can be accessed at time $t$, which is the focus of this work.
The space and time complexity of self-attention is linearly correlated to the size of this set $|C_t|$, making it an important metric of efficiency. 
For the rest of the paper, we will refer to $|C_t|$ as the \emph{memory size}.

Including all previous timesteps in self-attention by setting $C_t=\{1, \dots, t-1\}$ results in a quadratic complexity $\mathcal{O}(T^2)$ to compute the full attention over a sequence of length $T$. \textit{Fixed-spans}~\citep{dai2019transformer} take a more scalable approach such that $C_t = \{t-L, \dots, t-1 \}$ so the attention is restricted to previous $L$ steps. The total complexity in this case is $\mathcal{O}(TL)$, where $L$ is the attention span.

\textit{Adaptive-Span}~\citep{sukhbaatar2019adaptive} further improves upon upon this by learning an optimal span $L$ per attention head from data, which results in small $L$ values for many  heads. 
\textit{Compression} approaches~\citep{rae2020compressive} reduce memory size by compressing multiple timesteps into a single memory, with complexity $\mathcal{O}(TL/c)$, where $c$ is the compression rate. However, in all these approaches, all memories are treated equally without regards to their importance to the task. In this work, we focus on distinguishing between relevant and irrelevant memories by learning to expire unneeded information --- by expiring, the remaining attention on relevant information can scale beyond existing long context memory approaches.

\section{Expire-Span}

We describe \textsc{Expire-Span} and how to integrate it into Transformers to focus on relevant information and expire the rest, meaning memories can be permanently deleted. We describe how to scale \textsc{Expire-Span} and practically train with drastically longer memory spans.\footnote{The full implementation can be found at \url{https://github.com/facebookresearch/transformer-sequential}.}

\begin{wrapfigure}{R}{0.2\textwidth}
    \begin{center}
    \vspace{-35pt}
     \scalebox{0.8}{\begin{tikzpicture}[
        thick,
      >=stealth',
    ]
      \coordinate (O) at (0,0);
      \fill[gray!40!white] (-1.5,0) -- (0,1) -- (1, 1) -- (1,0) -- cycle;
      \draw[dashed,->] (-1.5,0) -- (1.5,0) coordinate[label = {below:$x$}] (xmax);
      \draw[dashed,->] (0,0) -- (0,1.5) coordinate[label = {left:$m(x)$}] (ymax);
      \draw (0,1) node[label = {left: $1$}] {};
      \draw[-] (0,1) -- (1,1) node[] {};
      \draw[-] (-1.5,0) -- (0,1) node[] {};
      \draw[-] (-2.5,0) -- (-1.5,0) node[] {};
      \draw (-1.5,0) node[label = {below: $-R$}] {};
\end{tikzpicture}    }  

    \end{center}
    \vspace{-10pt}
    \caption{Soft Mask}
    \vspace{-5pt}
    \label{fig:ramp}
\end{wrapfigure}

\subsection{Method}

\textsc{Expire-Span}, depicted in~\fig{model}, allows models to selectively forget memories that are no longer relevant.
We describe it in the context of a single Transformer layer and omit the layer index $l$ for brevity.
Our goal is to reduce the size of $C_t$ defined in~\secc{back} for more efficiency without performance degradation. For each memory $\vec{h}_i \in \mathbb{R}^d $, we will compute a scalar \textsc{Expire-Span} $e_i \in [0, L]$:
\begin{equation}
e_i = L \sigma(\vec{w}^{\top}\vec{h}_i + b).
\label{eq:span}
\end{equation}
Here $\vec{w} \in \mathbb{R}^d$ and $b \in \mathbb{R}$ represent trainable parameters, $\sigma$ is the sigmoid function, and $L$ is the maximum span. 
This expire-span $e_i$ determines how long $\vec{h}_i$ should be kept and included in $C_t$.
At time $t$, the remaining span of $\vec{h}_i$ is $r_{ti} = e_i - (t-i)$.
When $r_{ti}$ becomes negative, it indicates the memory $\vec{h}_i$ is expired and can be removed from $C_t$.
This can be implemented by updating attention weights $a_{ti}$ with a binary masking function $m_{ti} = \mathbf{1}_{r_{ti} > 0}$:

\begin{align}
a'_{ti} = \frac{m_{ti} a_{ti}}{\sum_j m_{tj} a_{tj}} , \quad \vec{o}_t = \sum_i a'_{ti} \vec{v}_i .
\end{align}

However, with such discrete masking, the Expire-Span $e_i$ will not receive any gradient for training. Instead, we use a soft masking function from~\citet{sukhbaatar2019adaptive} that smoothly transitions from 0 to 1 (see \fig{ramp}):
\begin{equation}
m_{ti} =  \max(0, \min(1, 1 + r_{ti}/R )),
\end{equation}
where $R$ is a hyperparameter that determines the length of a ramp that is bounded between 0 to 1. This function has non-zero gradient for values in $[-R, 0]$ to train $e_i$, but also can take a value of 0, which is necessary for expiring memories.
Thus $C_t = \{i \mid m_{ti} > 0\}$. Since $m_{ti}$ is a monotonically decreasing function of $t$, once a memory is expired, it can be permanently deleted.

Our goal is to reduce the average memory size, which is directly related with the average \textsc{Expire-Span}: 
\begin{align}
\frac{1}{T}\sum_t |C_t| 
&= \frac{1}{T}\sum_t \sum_{i<t} \mathbf{1}_{m_{ti} > 0} \nonumber \\
&= \frac{1}{T}\sum_i \left(R + \sum_{t>i}  \mathbf{1}_{r_{ti} > 0}\right) \nonumber \\
&= \frac{1}{T}\sum_i  \left(R + \sum_{t>i} \mathbf{1}_{e_i > t-i}\right) \nonumber \\
&= R - 1 + \frac{1}{T}\sum_i \left\lfloor e_i \right\rfloor
\end{align}
Therefore, we add an auxiliary term to the loss function to penalize the L1-norm of \textsc{Expire-Span}: 
\begin{equation}
L_\text{total} = L_\text{task} + \alpha \sum_i e_i/T,
\end{equation}
where $\alpha>0$ is a hyperparameter. This term decreases the span of memories that contribute less to the main task, resulting in a small memory that focuses only on relevant information. Note the new parameters, $\vec{w}$ and $b$, and the computations of \textsc{Expire-Spans} are negligible in size compared to the total number of parameters and computations.

\subsection{Adding Expire-Span to Transformers}
We describe how \textsc{Expire-Span} can be utilized within Transformer self-attention layers to decrease the memory size and focus on salient information. This section describes each modification clearly, to facilitate easier reproduction. We discuss practical training concerns, such as efficiency and regularization. 
Additional details can be found in the appendix.

\paragraph{Modifications to Multi-Head Attention}
Self-attention consists of multiple heads that have different keys, values, and queries. However, they all share one underlying memory, so a memory cannot be removed if it is used by any of the heads. Thus, we compute an \textsc{Expire-Span} at each layer that is shared amongst the heads.

\paragraph{Block Parallel}

This modification allows memories to be permanently deleted in \textsc{Expire-Span}. We use the caching mechanism~\citep{dai2019transformer}, 
where a block of timesteps $\mathrm{B}=[t, \dots, t+K-1]$ is processed in parallel for efficiency --- once a block is computed, its hidden states $[\vec{h}_t, \dots, \vec{h}_{t+K-1}]$ are cached so that future blocks can attend to them. 
This means a memory can be deleted only if it is not used by any of the queries in $\mathrm{B}$.
Concretely, $\vec{h}_i$ will be deleted when $m_{ti} = 0$ where $t$ is the first token of $\mathrm{B}$. However, this is not a concern for very long-term memories where $L \gg K$. 

\paragraph{Loss Computation}
The L1-norm loss for \textsc{Expire-Span} must be computed for every memory $\vec{h}_i$. A straightforward way is to compute it for the current block $\mathrm{B}$.
This empirically results in poor performance --- a possible explanation is that the time between positive and negative gradients on $e_i$ may become too distant. Negative gradients that increase $e_i$ only come from the main loss $L_\text{task}$ through the masking function $m_{ti}$, which has non-zero gradients only when memory $\vec{h}_i$ is about to expire with $0 < m_{ti} < 1$ for $t \in \mathrm{B}$. For a large $L \gg K$, $\vec{h}_i$ may have been computed many blocks before and since then the model weights would have changed. In contrast, the positive gradients that decrease $e_i$ are computed on the current block $i \in \mathrm{B}$.
To remove this discrepancy, we compute the auxiliary loss on $e_i$ at the same time as negative gradients when $0 < m_{ti} < 1$ for $t \in \mathrm{B}$.

\paragraph{Regularization}

A potential challenge in exceptionally long memory is greater capacity to overfit. As \textsc{Expire-Span} can scale to memories in the tens of thousands, it can overfit to learning specific span sizes on the training set that do not generalize. 
As a form of regularization, we propose to randomly shorten the memory during training. For each batch, we sample $l \sim \mathcal{U}(0,L)$ and set $a_{ti}=0$ for all $t-i > l$ only during training. This way, the model cannot assume the memory will always contain specific information, as the memory is randomly shortened.

\paragraph{Stable Training with Extremely Large Spans}
Multiplier $L$ in \eq{span} is the maximum span, so it can take very large values, exceeding tens of thousands. This is a potential problem because small changes in $\vec{h}_i$ or $\vec{w}$ will be amplified in \textsc{Expire-Span} $e_i$, and subsequently have dramatic effects on the model behaviour. As a straightforward remedy, for very large $L$ values, we replace \eq{span} with
\begin{equation}
e_i = L \sigma\left( (\vec{w}^{\top}\vec{h}_i + b ) / R \right) .
\end{equation}

\section{Experiments and Results}

\begin{figure*}[t]
    \centering
    \includegraphics[width=1\linewidth]{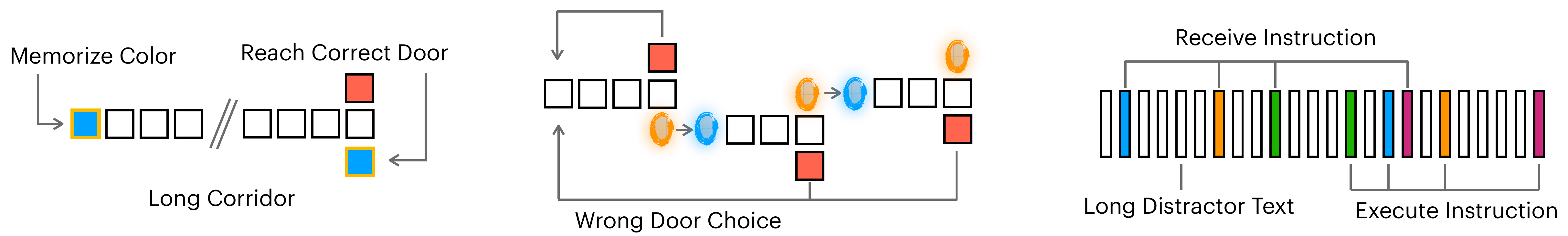}
    \caption{\textbf{Corridor Task (left)}- Agents must memorize the color of an object and walk through the door of the corresponding color at the end of a long corridor. \textbf{Portal Task (middle)}- An agent must trial-and-error to memorize the  sequence of doors. \textbf{Instruction Task (right)}- A model must recognize instructions, memorize them, and execute when at the correct location.}
    \label{fig:corridor_img}
\end{figure*}

We show that \textsc{Expire-Span} focuses on salient information on various constructed and real-world tasks that necessitate expiration. First, we describe baselines and efficiency metrics for comparing various models. Second, we illustrate the importance of expiration on various constructed tasks. Then, we highlight the scalability of \textsc{Expire-Span} when operating on extremely large memories.
Additional experiments and details are in the appendix.

\begin{figure*}[t]
    \includegraphics[width=0.31\linewidth]{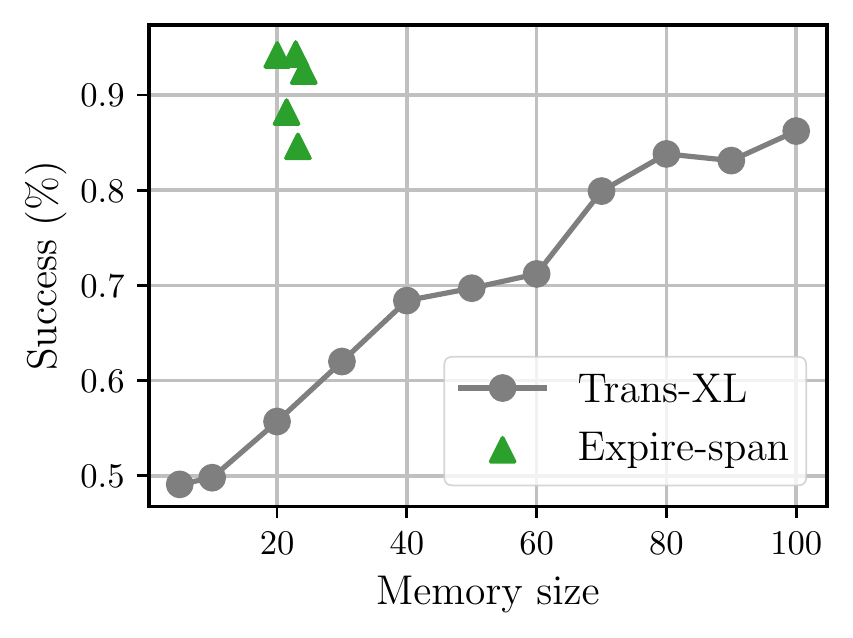}
    \hspace{0.3cm}
    \includegraphics[width=0.305\linewidth]{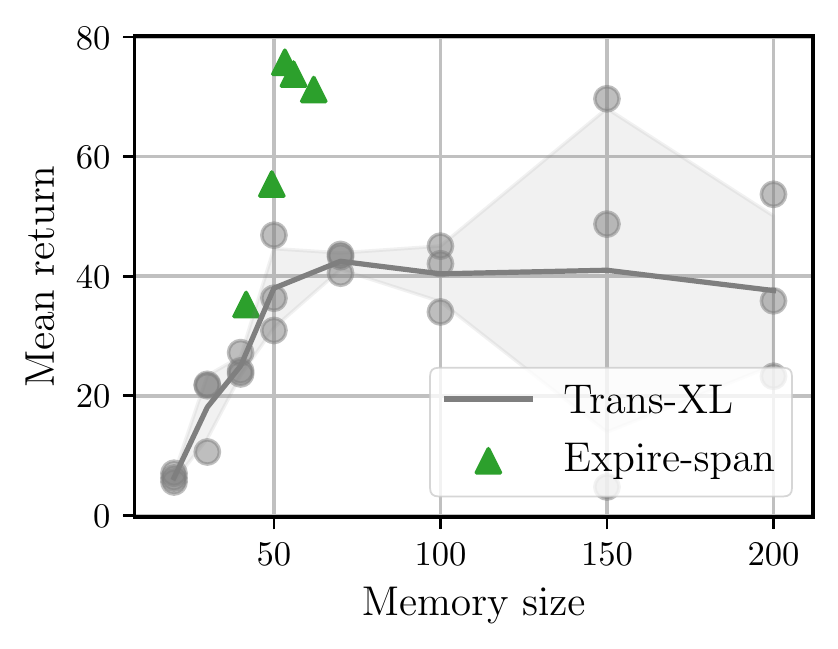}
    \hspace{0.3cm}
    \includegraphics[width=0.31\linewidth]{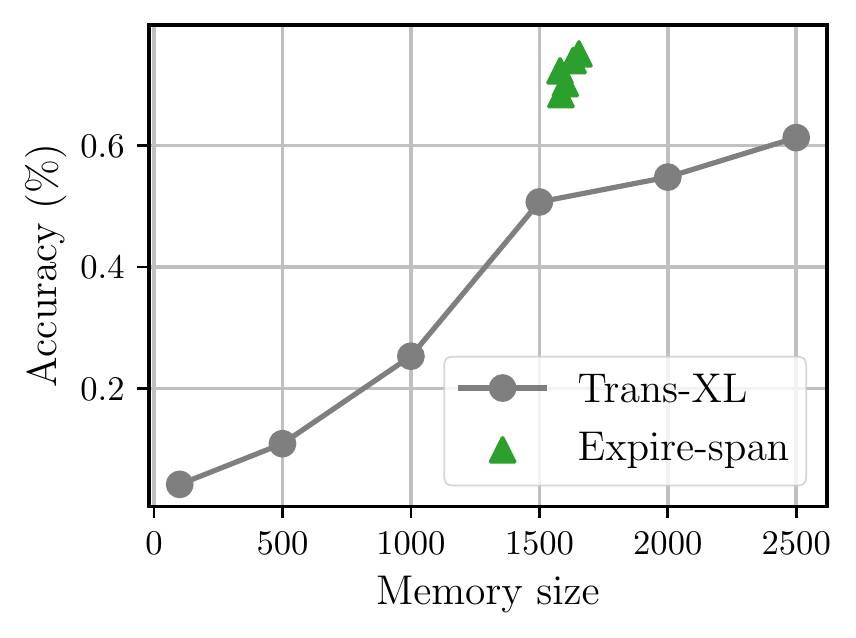}
    \hspace{0.3cm}
    \vspace{-3mm}
    \caption{We plot performance as a function of memory size for three tasks. Training scores are shown. Ideal models can achieve strong performance with small memories by identifying which information is important to remember. \textbf{Corridor Task (left)} --- We train 10 baseline models with different memory sizes, and five \textsc{Expire-Span} models with different seeds. 
    \textbf{Portal Task (middle)}- We train models with different memory sizes and random seeds. \textbf{Instruction Task (right)} --- We train 6 baseline models with different memory sizes, and five \textsc{Expire-Span} models with different seeds.}
    \label{fig:corridor_result}
\end{figure*}

\subsection{Baselines}
We compare our method against several baselines from~\secc{back} that take different approaches to access information in the past. 
We compare the performance of these methods, along with two efficiency metrics: GPU memory and training speed for a fixed model size and batch size.
First, we compare to \textit{Transformer-XL}~\citep{dai2019transformer}, which corresponds to the fixed-span approach where simply the last $L$ memories are kept. 
Our Transformer-XL implementation also serves as a base model for all the other baselines to guarantee that the only difference among them is how memories are handled. 
The other baselines are \textit{Adaptive-Span}~\citep{sukhbaatar2019adaptive} and \textit{Compressive Transformer}~\citep{rae2020compressive}, two popular approaches for long memory tasks. For Compressive Transformer, we implemented the  mean-pooling version, which was shown to have strong performance despite its simplicity. 

\subsection{Importance of Expiration: Illustrative Tasks}

\paragraph{Remembering One Key Piece of Information}

To illustrate a case where proper expiration of unnecessary memories is critical, we begin with an RL gridworld task: walking down a corridor. In this \textit{Corridor} task, depicted in Figure~\ref{fig:corridor_img} (left), the agent is placed at one end of a very long corridor, next to an object that is either red or blue. The agent must walk down the corridor and go to the door that corresponds to the color of the object that it saw at the beginning to receive $+1$ reward. The requirement on the memory is very low: the agent only needs to remember the object color so it can walk through the correct door. 

\textsc{Expire-Span} models can take advantage of this fact and keep the memory size small regardless of the corridor length, which can vary between 3 and 200. 
This is confirmed in the results shown in Figure~\ref{fig:corridor_result} (left) where the \textsc{Expire-Span} models achieve high performance on this task with very small memories. 
Without the ability to forget, the Transformer-XL models require large memory for storing all navigation steps that grow with the corridor length.

\paragraph{Remembering Long Sequences of Information} 

Next, we analyze \textsc{Expire-Span} on another reinforcement learning task, but this time testing memorization of sequences: Portal through Multiple Rooms. An agent in a gridworld must navigate through multiple rooms separated by different doors, depicted in Figure~\ref{fig:corridor_img} (middle). Each room has two exit doors with different colors --- one door portals to the adjacent room, while the other portals back to the start. 
However, which door works in which room is random for each episode. 
Thus, the only way to visit more rooms is by trial-and-error, where agents need to remember the sequence of correct doors to successfully navigate to the end. 
The environment is  partially observable and randomized at each episode.

We display results in Figure~\ref{fig:corridor_result} (middle). The Transformer-XL models need longer memory to perform better and visit more rooms, because each new room requires many navigation steps to reach.
However, those navigation steps are actually irrelevant because the agent only needs to memorize the colors of the correct doors.
Usually, the agent needs to pass through the same room multiple times to solve the remaining rooms, but it only needs to remember the door color from the first pass, while all subsequent passes can be expired.
Since \textsc{Expire-Span} models can discard irrelevant memories and focus its memory on memorizing the exact sequence of door colors, they achieve strong performance with much smaller memory compared to the Transformer-XL baseline.

\paragraph{Remembering Long Sequences with Severe Distractors}

To illustrate a more difficult task where a model must learn to expire,
we use a dialogue-based story generation task from the LIGHT~\citep{urbanek2019learning} text world game environment. The model visits various locations and \textit{receives} instructions of the form \textit{can you tell the [butler] that the [town official] wants to see them?}. When the model is in a location where the \textit{butler} is present, they must \textit{execute} the instruction by generating \textit{You tell the butler ``town official wants to see you!''}.  Between receiving and executing, thousands of words of distractor text exist as shown \fig{corridor_img} (right). The model must learn to expire the distractors. Note multiple instructions can be in queue for execution.

We experiment with a dataset where the average distance between receiving and executing instructions is around 950 distractor words. Models are trained as language models, but evaluated only on their success in executing the instruction. 
Task details and model architecture are provided in the appendix. We illustrate in Figure~\ref{fig:corridor_result} (right) that \textsc{Expire-Span} is much more successful at this task than Transformer-XL and Adaptive-Span (see the appendix), as it can focus on the specific instruction lines.

\subsection{Scalability of Expire-Span}

We analyze the scalability of \textsc{Expire-Span}. On a copy task, we train models with spans up to 128k timesteps. Then, we show the utility of \textsc{Expire-Span} on character-level language modeling --- Enwik8 and PG-19 --- and a moving objects task that is processed frame by frame. For these tasks, we also analyze the efficiency of \textsc{Expire-Span} compared to existing methods, and demonstrate that our method has a smaller memory footprint and faster processing speed. We quantify efficiency with two metrics: \textbf{(1)} peak GPU memory usage and \textbf{(2)} training time per batch (comparing fixed batch size for similar size models).

\paragraph{Extremely Long Copy}

To illustrate the scalability of \textsc{Expire-Span}, we construct a copy task where the model sees a sequence of \textit{A} very far in the past. The rest of the characters are \textit{B}. The model must copy the correct quantity of \textit{A}. We design the task such that a long span (up to 128k) can be required, as the \textit{A} tokens are very far into the past. In Table~\ref{tab:copy}, we show that only by scaling the maximum span to 128k it is possible to achieve improved performance. We compare to a Transformer-XL baseline with 2k attention span and a \textsc{Expire-Span} model with smaller span.

\begin{table}
    \centering
    \begin{tabular}{lcc}
    \toprule
    Model & Maximum span & Accuracy (\%) \\
    \midrule
    Transformer-XL & 2k &  26.7 \\ 
    \textsc{Expire-Span} & 16k &  29.4 \\ 
    \textsc{Expire-Span} & 128k & \bf 52.1 \\ 
    \bottomrule
    \end{tabular}
    \caption{
    \textbf{Copy Task.} We report accuracy on the test set.
    }
    \label{tab:copy}
\end{table}

\paragraph{Character Level Language Modeling: Enwik8} 

\begin{figure}[t]
    \centering
    \includegraphics[width=0.8\linewidth]{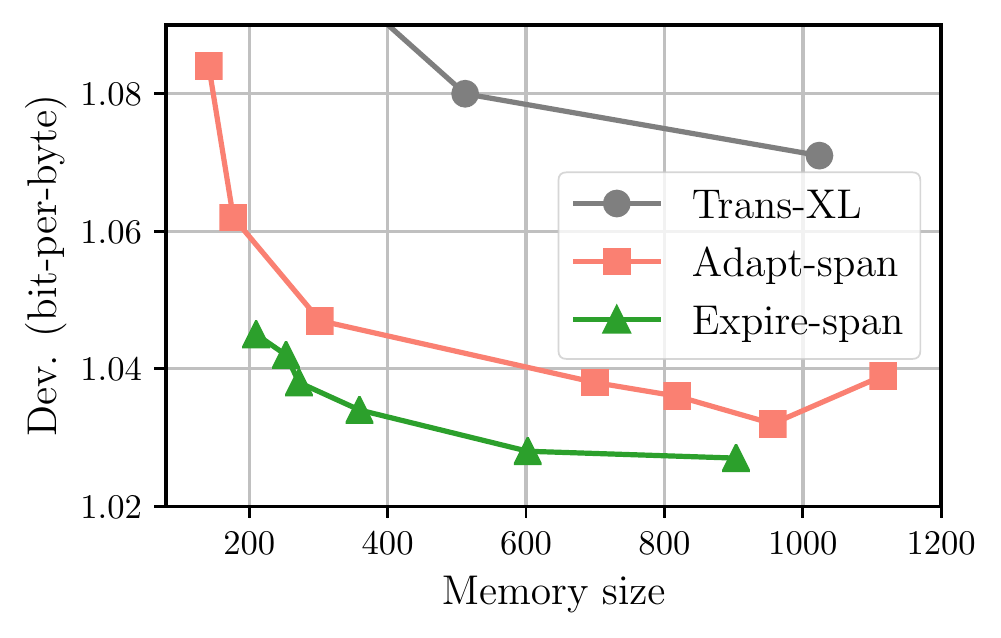}
    \caption{\textbf{Performance as a Function of Memory Size on Enwik8.} Lower bpb and smaller memory size is better.}
    \label{fig:enwik8_span}
\end{figure}

\begin{table}[t]
\setlength{\tabcolsep}{2.0pt}
    \centering
    \begin{tabular}{lcc}
    \toprule
    Model & Params & Test \\
    \midrule
    \multicolumn{3}{l}{\small\emph{Small models}} \\
    Trans-XL 12L~{\footnotesize {\citep{dai2019transformer}}} & 41M  & 1.06 \\
    Adapt-Span 12L~{\footnotesize {\citep{sukhbaatar2019adaptive}}} & 39M & 1.02 \\
    Our Trans-XL 12L baseline & 38M  & 1.06 \\
    \textsc{Expire-Span} 12L & 38M & \bf 0.99 \\ 
    \midrule
    Trans-XL 24L~{\footnotesize {\citep{dai2019transformer}}} & 277M  & 0.99 \\
    Sparse Trans.~{\footnotesize {\citep{child2019generating}}} & 95M  & 0.99 \\
    Adapt-Span 24L~{\footnotesize {\citep{sukhbaatar2019adaptive}}} & 209M & 0.98 \\
    All-Attention~{\footnotesize {\citep{sukhbaatar2019augmenting}}} & 114M  & 0.98 \\
    Compressive Trans.~{\footnotesize {\citep{rae2020compressive}}} & 277M  & 0.97 \\
    Routing Trans.~{\footnotesize {\citep{roy2020efficient}}}& - & 0.99 \\ 
    Feedback Trans.~{\footnotesize {\citep{fan2020accessing}}} & 77M &  0.96 \\    
    \textsc{Expire-Span} 24L & 208M &  \bf 0.95 \\
    \bottomrule
    \end{tabular}
    \caption{
    \textbf{Enwik8 Results.} We report bit-per-byte (bpb) on test and the number of parameters.
    }
    \label{tab:enwik8}
\end{table}

\begin{figure}
    \centering
    \includegraphics[width=\linewidth]{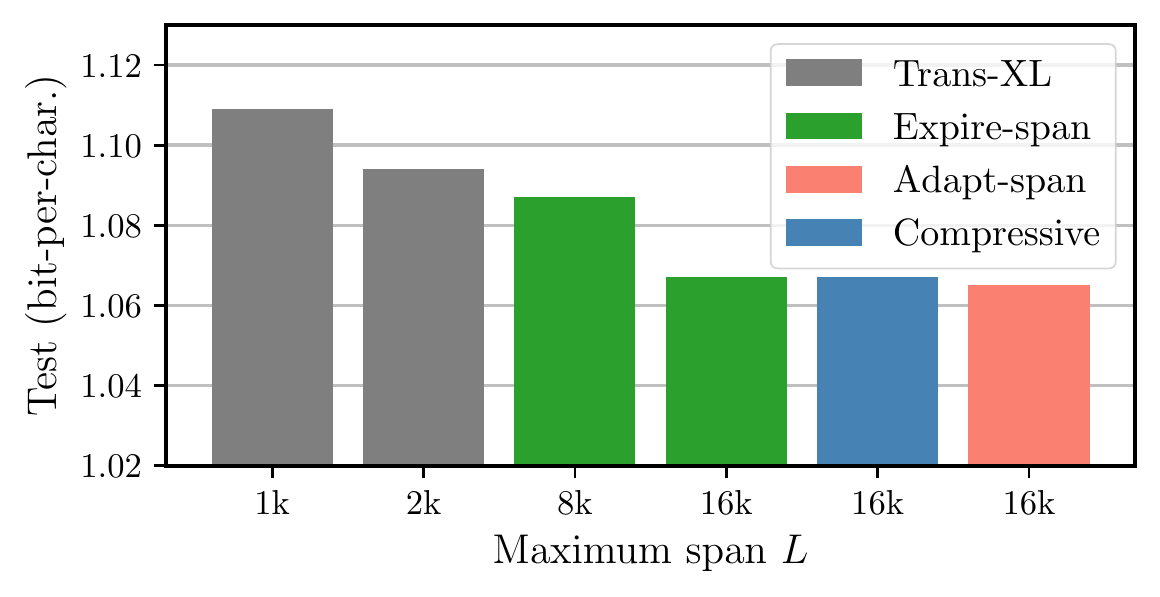}
    \caption{\textbf{Performance on Character-level PG-19.}  We report bit-per-character on test.}
    \label{fig:pg19}
\end{figure}

We subsequently experiment on Enwik8 for character level language modeling~\citep{mahoney2011large}.  
We compare the performance of \textsc{Expire-Span} with Adaptive-Span and Transformer-XL, varying the average span size (see \fig{enwik8_span}). Models with \textsc{Expire-Span} achieve stronger results --- when comparing at any given memory size, \textsc{Expire-Span} outperforms both baselines. Further, the performance of \textsc{Expire-Span} does not vary much even if the memory size is drastically reduced, indicating the model retains a small quantity of salient information for good performance.

Next, we compare \textsc{Expire-Span} to existing work in \tab{enwik8}. A small \textsc{Expire-Span} model with the maximum span $L=16\text{k}$ outperforms similarly sized baselines by a large margin.
We also trained a larger \textsc{Expire-Span} model with $L=32\text{k}$ and LayerDrop~\cite{Fan2020Reducing}, which outperforms the Compressive Transformer and sets a new state of the art on this task. 
This indicates that models can learn to expire relevant information and encode long context effectively, even on very competitive language modeling benchmarks.

Finally, we compare the efficiency of \textsc{Expire-Span} with the Transformer-XL, Adaptive-Span and Compressive Transformer baselines. 
We find that \textsc{Expire-Span} models achieve much better performance, as shown in Table~\ref{tab:efficiency} with substantially less GPU memory and faster training time per batch.

\paragraph{Character Level Language Modeling: PG-19} 

We use the PG-19~\citep{rae2020compressive} benchmark and convert it to character-level language modeling with a vocabulary size of 3506.
We train several baselines: Transformer-XL with maximum spans of 1k and 2k, and Adaptive-Span and Compressive Transformers with 16k span.
We train \textsc{Expire-Span} with maximum spans of 8k and 16k.
We present results in Figure~\ref{fig:pg19}, where we show that  \textsc{Expire-Span} is substantially better than Transformer-XL, and matches the performance of Adaptive-Span and Compressive Transformer.

\begin{figure}[t]
    \centering 
    \includegraphics[width=0.8\linewidth]{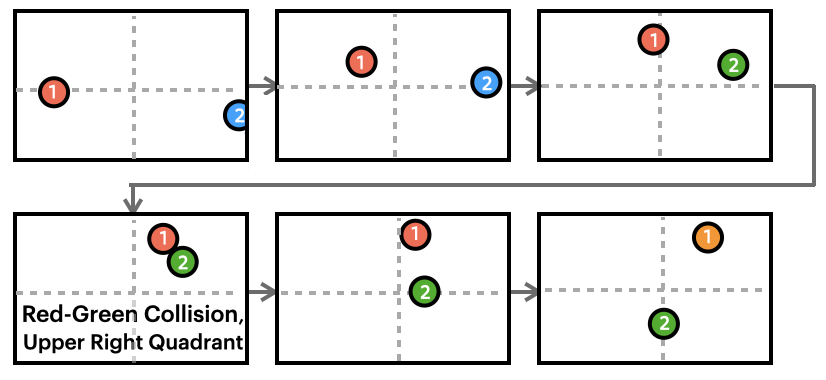}
     \caption{\textbf{Object Collision} task tests if models can remember the location of specified colored collisions.}
    \label{fig:collision}
\end{figure}

However, \textsc{Expire-Span} uses its available memory very effectively. The 16k maximum span \textsc{Expire-Span} model has an average memory size of 860. 
In comparison, the Adaptive-Span model has an average memory size of 2440, almost 3x that of the 16k \textsc{Expire-Span} model. This indicates that \textsc{Expire-Span} enables models to identify the critical bits of information and expire the rest, reaching the same performance with a much smaller memory.

Finally, comparing efficiency (Table~\ref{tab:efficiency}),  \textsc{Expire-Span} trains at double the speed of Compressive Transformer. \textsc{Expire-Span} is faster than Adaptive-Span, though uses slightly more memory. The memory usage of \textsc{Expire-Span} is usually lower, around 12GB, but spikes for some sentences. Lastly, while the average span size of \textsc{Expire-Span} is lower than Adaptive-Span, the computation requires additional tensors allocated in memory, which can potentially be addressed by an optimized implementation. 

\paragraph{Frame-by-Frame Processing: Object Collision} 

An important setting where learning which long context may be important is in video understanding, a field with increasing focus as model architectures provide the capability to process long sequences.  Despite video data being memory intensive, salient events might be localized in space and time.
We test our model on a task where two objects move around and collide, and the goal is to reason about the location of specified-color collisions. Objects have a color that can randomly change. We divide the grid into four quadrants and the model is asked to recall the quadrants of the last collision of a specific color pair. Because the collisions are rare, and collisions of specific colors are even rarer, the model must process a large quantity of frames.

\begin{table}[t]
    \centering
    \begin{tabular}{lcc}
    \toprule
    Model & Maximum Span & Test Error (\%) \\
    \midrule
    Transformer-XL & 1k & 73.3 \\
    Compressive & 8k & 63.8 \\
    Adaptive-Span & 16k & 59.8 \\
    \midrule 
    \multirow{3}{*}{\textsc{Expire-Span}}  & 16k & 52.2 \\
     & 32k & 36.7 \\ 
     & 64k & \bf 26.7 \\ 
    \bottomrule
    \end{tabular}
    \caption{
    \textbf{Results on Object Collision}. We report the error on the test set comparing to various baselines.
    }
    \label{tab:colliding}
\end{table}

\begin{table*}[t]
    \centering
    \renewcommand{\arraystretch}{0.8}
    \begin{tabular}{lllccc}
    \toprule
    && Model &  Performance & GPU Memory (GB) & Time/Batch (ms) \\
    \midrule
    \multirow{4}{*}{Enwik8} && Transformer-XL  & 1.06 bpb & 27 & 649 \\ 
    && Compressive Transformer  & 1.05 bpb & 21 & 838 \\ 
    && Adaptive-Span & 1.04 bpb & 20 & 483 \\ 
    && \textsc{Expire-Span} & {\bf  1.03} bpb & \bf  15 & \bf 408 \\ 
    \midrule 
    \multirow{3}{*}{Char-level PG-19} && Compressive Transformer  & 1.07 bpc & 17 & 753 \\ 
    && Adaptive-Span & 1.07 bpc & \bf 13 & 427 \\ 
    && \textsc{Expire-Span} &  1.07 bpc & 15 & \bf 388 \\ 
    \midrule 
    \multirow{3}{*}{Object Collision} && Compressive Transformer  & 63.8\% Error & \bf 12 & 327 \\ 
    && Adaptive-Span & 59.8\% Error & 17 & 365 \\ 
    && \textsc{Expire-Span} & {\bf  52.2\%} Error & \bf  12 & \bf  130 \\ 
    \bottomrule
    \end{tabular}
    \caption{
    \textbf{Efficiency of \textsc{Expire-Span}}. We report peak GPU memory usage and per-batch training time, fixing the batch size. 
    }
    \label{tab:efficiency}
    \vspace{-2mm}
\end{table*}

We illustrate the task in Figure~\ref{fig:collision} and results in Table~\ref{tab:colliding}. The task requires many frames, so long context is very beneficial --- as the \textsc{Expire-Span} maximal span increases, performance steadily rises. Our largest span, 64k, matches the size of the largest attention limit reported to date~\citep{kitaev2019reformer} and has the strongest performance. This model is trained with the random drop regularization method described in Section 4.2. Compared to Compressive Transformer and Adaptive-Span baselines, our \textsc{Expire-Span} model has the strongest performance. 

Comparing efficiency, \textsc{Expire-Span} trains almost 3x faster than both baselines (see Table~\ref{tab:efficiency}) while having much stronger performance. Further, expiration is critical to this performance --- a Adaptive-Span model with $L=32\text{k}$ runs out of memory in the same setting where we trained our \textsc{Expire-Span} model with $L=64\text{k}$. Through expiration, our model can keep the GPU memory usage reasonable and train with the longer spans necessary for strong performance.

\begin{figure*}[t]
\begin{minipage}{0.63\textwidth}
    \centering 
    \includegraphics[width=\linewidth]{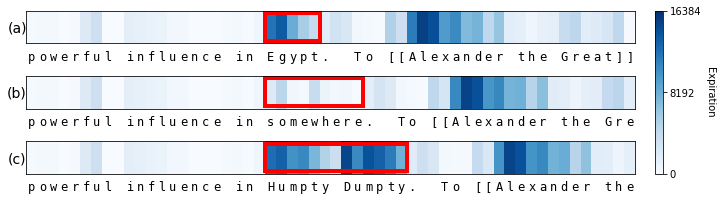}
    \caption{\textbf{Expiration in \textsc{Expire-Span} on Enwik8}. In \textbf{(a)}, the model strongly memorizes two areas,  ``Egypt'' and ``Alexander''. In \textbf{(b)}, if we replace ``Egypt'' with ``somewhere'', then it's forgotten fast. In \textbf{(c)}, we insert ``Humpty Dumpty'' and the model retains these rare words in memory.}
    \label{fig:vis_enwiki8}
\end{minipage}
\hfill
\begin{minipage}{0.33\textwidth}
\setlength{\tabcolsep}{3.2pt}
    \includegraphics[width=0.99\linewidth]{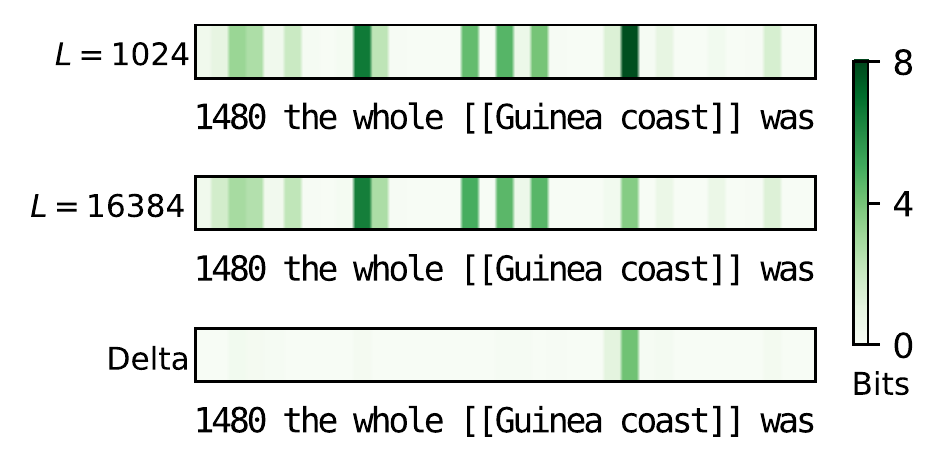}
    \caption{\textbf{Accuracy Needs Memory.} As the maximum span is artificially decreased at inference time from 16k to only 1k, the prediction is less accurate.}
    \label{fig:enwik8_loss}
\end{minipage}
\end{figure*}

\section{Analysis and Discussion} 

\textsc{Expire-Span} creates the phenomena of \textit{selective forgetting}: it allows memories to be permanently deleted if the model learns they are not useful for the final task. In this section, we analyze the information retained and expired by \textsc{Expire-Span} models to better understand how models use the ability to forget. Additional analyses are in the appendix.

\paragraph{Retaining Salient Information}

We analyze what is retained by an \textsc{Expire-Span} model on Enwik8 to understand how models utilize the ability to forget. In Figure~\ref{fig:vis_enwiki8} (a), we show that the model retains information about named entities such as \textit{Egypt} and \textit{Alexander the Great} by giving them longer spans (darker color). Next, we analyze how expire-spans changes when we artificially edit the past text. In Figure~\ref{fig:vis_enwiki8} (b), we replace the entity \textit{Egypt} with the generic text \textit{somewhere}, and this generic word is quickly expired. In Figure~\ref{fig:vis_enwiki8} (c), we edit \textit{Egypt} to \textit{Humpty Dumpty}, which is a very rare entity, and the model retains it in memory without expiring.
In addition to entities, \textsc{Expire-Span} memorizes spaces, newlines, and section titles, all of which retain information about words, sentences, or sections. The model's expiration choices vary by layer, indicating that \textsc{Expire-Span} models use the memory at each layer to remember different information.

\paragraph{Importance of Long Term Memory}
Next, we analyze which predictions benefit the most from memory capacity. We take an \textsc{Expire-Span} model trained on Enwik8 and decrease the maximum span size to 1024 at inference time, even though the model was trained with a maximum span of 16k. We then compare which predictions decreased in accuracy. In Figure~\ref{fig:enwik8_loss}, we see that  models have a much higher loss when predicting the named entity \textit{Guinea coast} compared to having the full 16k maximal span. \textit{Guinea coast} was mentioned 3584 tokens earlier, which indicates that long attention is often necessary to predict words mentioned in far away context. In general, we found that rare tokens and structural information about documents, such as section headings or document titles, required longer attention span to accurately predict.

\paragraph{Efficiency Advantages of Expire-Span} 

Finally, we end with a brief discussion about why \textsc{Expire-Span} is more efficient compared to existing architectures that focus on long context. First, Transformer-XL cannot adapt to the data at all, so it becomes slow and inefficient quite quickly as the span size increases. Adaptive-Span can adapt to the data and adjust its memory, but this memory size is fixed after training and does not have the dynamic adjustment of Expire-Span (where memory depends on local context even at inference time). Finally, the Compressive Transformer compresses past memories, but it compresses always at a fixed rate. The compression rate is an adjustable parameter, but aggressive compression potentially hurts performance. In contrast, \textsc{Expire-Span} can expire irrelevant content, which both improves performance by focusing on salient information, and reduces the load on GPU memory and allows for faster processing per batch.

\vspace{-2mm}
\section{Conclusion}
\vspace{-2mm}
We present \textsc{Expire-Span}, an operation that can be added to any attention mechanism to enable models to learn what to forget. By expiring irrelevant information, models can scale attention to tens of thousands of past memories. We highlight the strong performance of \textsc{Expire-Span} in language modeling,  reinforcement learning, object collision, and algorithmic tasks, and use it to attend over tens of thousands of past memories. The scalability and much greater efficiency of our proposed \textsc{Expire-Span} method has strong potential for allowing models to be applied to more challenging, human-like tasks that would require expiration.

\bibliography{iclr2021_conference}
\bibliographystyle{icml2021}

\clearpage 
\newpage 
\appendix
\section{Appendix}

\subsection{Additional Method Details}
\paragraph{Position Embedding}
Relative position embeddings~\citep{shaw2018self} make it possible to condition on the order of inputs by modifying the attention to $a_{ti} = \text{Softmax}(\vec{q}_t^\top \vec{k}_i + \vec{q}_t^\top \vec{p}_{t-i})$.
However, because this second term is computed for the whole block in parallel for efficiency, 
it can become expensive for a large $L$ even when the average memory size $|C_t|$ is small. 
Our solution is to remove position embeddings from older memories $i<t-K$ (where $K$ is the block size), which empirically does not affect performance. The computational complexity of the position embeddings is then $\mathcal{O}(K)$, thus allowing us to increase the maximum span $L$. This modification makes training \textsc{Expire-Span} more efficient, but does not improve accuracy.

\paragraph{Training with Small Initial Spans}
\textsc{Expire-Span} scales to long attention spans as it quickly learns to expire irrelevant content. However, at the beginning of training, the long span can use large quantities of GPU memory. To circumvent this, we initialize the bias term $b$ with a negative value.
This prevents large memory usage at the beginning of training, after which the model quickly learns to expire and the memory usage is no longer problematic.

\subsection{Additional Experimental Results}
\paragraph{Efficiency for Instruction Task}

We include a comparison of \textsc{Expire-Span} to Adaptive-Span and Compressive Transformer in Table~\ref{tab:efficiency_appendix} and show that \textsc{Expire-Span} has stronger performance, is faster, and saves GPU memory. 

\begin{table*}[t]
    \centering
    \renewcommand{\arraystretch}{0.8}
    \begin{tabular}{lllccc}
    \toprule
    && Model &  Performance & GPU Memory (GB) & Time/Batch (ms) \\
    \midrule
    \multirow{3}{*}{Instruction Task} && Compressive Transformer & 71\% Acc & 10 & 210 \\ 
    && Adaptive-Span & 64\% Acc & 14 & 240 \\ 
    && \textsc{Expire-Span} & {\bf  74\%} Acc & \bf  8 & \bf  90 \\ 
    \bottomrule
    \end{tabular}
    \caption{
    \textbf{Efficiency of \textsc{Expire-Span}}. We report peak GPU memory usage and per-batch training time, fixing the batch size. We evaluate the mean pooling version of the Compressive Transformer. 
    }
    \label{tab:efficiency_appendix}
    \vspace{-2mm}
\end{table*}

\paragraph{Wikitext-103 Language Modeling} 

\begin{table}[t]
    \centering
    \begin{tabular}{lccc}
    \toprule
    Model & Params & Test\\
    \midrule
	DEQ-Trans.~\citep{bai2019deep} & 110M & 23.3\\
    Trans-XL~\citep{dai2019transformer} & 257M  & 18.3  \\
    Feedback Trans.~\citep{fan2020accessing} & 77M & 18.3 \\    
    Trans.+LayerDrop~\citep{Fan2020Reducing} & 423M  & 17.7 \\
    Compressive Trans.~\citep{rae2020compressive} & 277M  & 17.1 \\
    Routing Trans.~\citep{roy2020efficient}& - &  15.8 \\ 
    \midrule 
    \textsc{Expire-Span} & 140M &  19.6 \\ 
    \bottomrule
    \end{tabular}
    \caption{
    \textbf{Wikitext-103 Results.} We report perplexity on test.
    }
    \label{tab:wiki103}
\end{table}

\begin{figure}[t]
    \centering
    \includegraphics[width=0.8\linewidth]{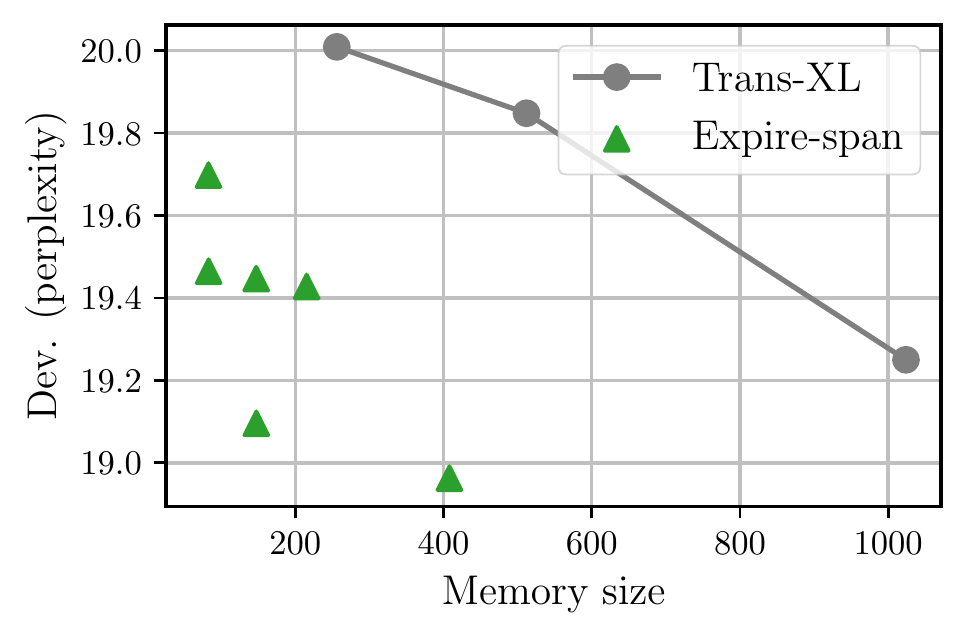}
    \caption{\textbf{Performance as a function of Memory Size on Wikitext-103}}
    \label{fig:wiki103_memsize}
\end{figure}

The Wikitext-103 word-level language modeling benchmark~\citep{merity2016pointer} consists of a collection of Wikipedia articles and a fixed vocabulary size of 270K.  We set the max attention span for \textsc{Expire-Span} to 8K. We compare \textsc{Expire-Span} to existing work in Table~\ref{tab:wiki103} and show that even fairly small models trained with \textsc{Expire-Span} achieve competitive results.
Next, we analyze the performance of \textsc{Expire-Span} on Wikitext-103 as the memory size increases. We compare to a Transformer-XL model in Figure~\ref{fig:wiki103_memsize} --- even with far smaller memory, \textsc{Expire-Span} performs much better.

\begin{figure*}[t]
    \centering
    {\includegraphics[width=\linewidth]{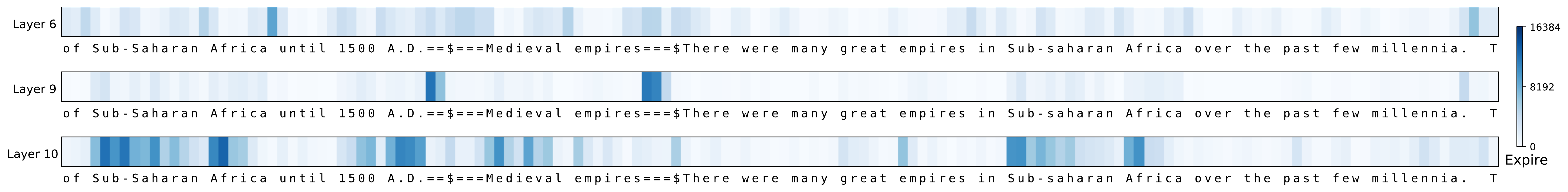}}
    \caption{\textbf{Per-Layer \textsc{Expire-Span} values on Enwik8}. We visualize the expire-spans of different layers: layer 6 gives long span to spaces, layer 9 memorizes special tokens like newlines and section titles, and layer 10 retains named entities in memory.} 
    \label{fig:vis_enwiki8_additional_layerwise}
\end{figure*}

\paragraph{Expire-span Performance and Analysis on Enwik8}
In Figure~\ref{fig:vis_enwiki8_additional_layerwise}, we analyze multiple layers of a trained model and show that different layers memorize different types of information. Several layers retain summarizing information about sentences or sections by increasing the expire-spans of spaces, new lines, and section titles. 

Additionally, we did an ablation by running our large Expire-Span model without LayerDrop. Its validation performance dropped from 0.98bpb to 1.00bpb.

\paragraph{Importance of Structured Dropout for Regularization}

\begin{figure*}
    \centering
    \includegraphics[width=4in]{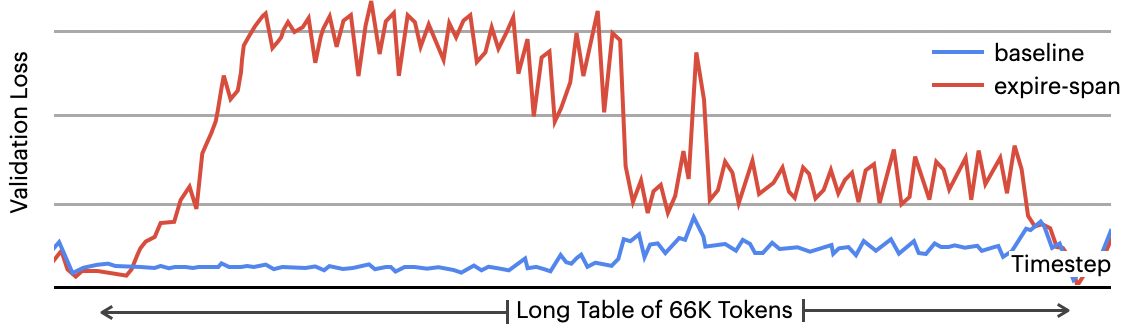}
    \caption{\textbf{Extreme Overfitting} on part of validation occurs without proper regularization.}
    \label{fig:enwik8_ovefit}
\end{figure*}

We analyze the importance of structured dropout to regularize the large memory capacity provided by \textsc{Expire-Span}. 
In an experiment on enwiki8, Figure~\ref{fig:enwik8_ovefit} shows that loss on a portion of  validation data was incredibly large. This part corresponds to a 66K token long table. We hypothesize that the model likely never encountered such a table during training. During validation, this caused all non-table tokens to expire. Without regularizing the model memory size during training, the model can easily overfit.

\paragraph{Colliding Objects, An Easy Version}
We experiment with an easier version of the Colliding Objects task where objects do not have colors. The model has to predict either the last collision, or a mapping of the last 3 collisions. In contrast to the harder task, there are no color switches and any collision prediction is valid. As this version is less memory intensive, the \textsc{Expire-span} model almost solves it with a shorter maximum span, as shown in \tab{colliding_easy}.

\begin{table}
    \centering
    \begin{tabular}{lcc}
    \toprule
    Model & Maximum Span & Test Error (\%)  \\
    \midrule
    Transformer-XL & 1k & 39.1 \\
    \midrule
     & 1k & 19.5 \\
    \textsc{Expire-Span} & 2k & 9.1 \\ 
     & 4k & 3.2 \\ 
    \bottomrule
    \end{tabular}
    \caption{
    \textbf{Colliding Objects Results}. We report test error. 
    }
    \label{tab:colliding_easy}
\end{table}

\subsection{Additional Implementation Details}

\subsubsection{Reinforcement Learning Tasks}
We used MazeBase~\citep{Sukhbaatar2015MazeBaseAS} to construct tasks in grid world.
Agents can observe its surrounding $3\times3$ area and  move in the four cardinal directions. Every objects and their properties are described by words such as ``agent'',``block'', ``blue'', etc.
Thus, the input to the model is a binary tensor of size $3\times3\times \text{vocabulary-size}$.

We train 2-layer Transformers with 64 hidden units using actor-Critic algorithm. We used a BPTT length of 100, and an entropy cost of 0.0005.

\paragraph{Corridor Task}
The corridor length is sampled from $\mathcal{U}(3, 200)$.
All models are trained for 100M steps.
We used RMSProp optimizer with a learning rate of 0.0001 and a batch size of 64.
For the expire-span models, we set the maximum span $L$ to 200, the loss coefficient $\alpha$ to 5e-6, and the ramp length $R$ to 16.

\paragraph{Multi-Room Portal}
In this task, there are 50 rooms sequentially connected together. Each room is $5\times5$ in size, and have two doors with different colors. If agent go to the correct door, it will be teleported to the next room, but if it is the wrong door, the agent will be teleported back to the first room and have to start over.
Which of the two doors is correct in each room is randomly decided and fixed throughout the episode.
This information is not visible to the agent, thus can only be discovered by trial and error within each episode.
The current room number is visible to the agent.

When the agent successfully transitions from the $k$-th room to the next, it receives a reward of $0.1k$. The episode ends if the agent makes two mistakes in the same room, reaches the last room, or when the number of steps reach 1000. A reward discount of 0.98 is used.
All models are trained with Adam optimizer with a learning rate of 5e-4, and a batch size of 1024, with gradients are clipped at 0.1. 
We set $L=100$, $R=16$ and $\alpha=$1e-6 for the expire-span models.

\subsubsection{Instruction Task in LIGHT}
We train 6-layer models with a hidden size of 512 and 8 attention heads. To train, we use the Adam optimizer with a learning rate of 7e-4 and 8000 warmup updates. We set the expire-span ramp $R$ to 64 and the expire-span loss $\alpha$ to 2e-6.

\subsubsection{Collision Task}
At the start of the simulation, each particle samples a Gaussian Normal velocity and position uniform inside a $16\times 16$ grid.  At each time step the particles' position is updated by adding its velocity (unless it would go off the grid, in which case its velocity is re-sampled).  
There are 5 different colors, and a particle can change its color randomly at each step with 0.05 probability.
A collision happens when the two particles have the same rasterized locations, but it does not affect the movement.

Given a question specifying two colors,
the task is to report in which of the four quadrants of the grid the last collision of the specified-colors occurred. To make the task easier to learn, 40\% of the queries will have the matching colors as the last collision.

The model is given an input sequence of tokens that has 8 entries per timestep.  The first 4 are the rounded and rasterized $(x,y)$ locations of the two particles, and next 2 are tokens representing the colors of the particles. The last 2 entries are ``question'' tokens that specify the colors of the collision.  The output sequence has a token for each quadrant.
We generate 50M steps for training, which equals to 400M entries.

\paragraph{Easy Version:}
The particles have no color in this version.
There are two types of questions, in which the task is to report either: \textbf{(i)} in which of the four quadrants of the grid the last collision occurred, or \textbf{(ii)} the label mapping of the last 3 collisions.

\subsubsection{Language Modeling Details} 

\paragraph{Enwik8}

Our small model has 12 layers with a hidden size of 512 and 8 attention heads.  To train, we use Adam optimizer with a learning rate of 7e-4, a batch size of 512, a block size of 512 and 8000 warmup updates. All models are trained for 100k updates. The model in Table~2 is further fine-tuned for another 10k updates with a 10x smaller LR. The baseline models used for comparison are the same size model following the same training protocol.

The large \textsc{Expire-Span} model Table~2 is a 24-layer model with a hidden size of 768 and 4096 feedforward units. It is trained with a learning rate of 4e-4 and 16k warmup steps. In addition to 0.5 dropout, we also add 0.2 layer-drop. The \textsc{Expire-Span} parameters are $L=32k$, $\alpha=$3e-7, and $R=128$. We used the version of Eq.~6 due to the very long maximum span.

\paragraph{Character-level PG-19}

Besides the maximum span, all model parameters and training parameters were held constant. Each model had 12 layers, a hidden size of 512, a feedforward size of 2048, 8 attention heads, and processed a block of 512 characters at a time. We initialized the weights using a uniform distribution as described by~\cite{glorot2010understanding}, used dropout of 0.2, clipped the gradients at 0.3, warmed up the learning rate linearly for 8000 steps, and used cosine annealing to decay the learning rate after warmup~\citep{loshchilov2016sgdr}. For the \textsc{Expire-Span} models, we used a ramp of $R=128$ and an expiration loss coefficient of $\alpha=$1e-6 (3e-7) for $L=8k$ ($16k$).

\paragraph{Wikitext-103}
All models have 8 layers and 1024 hidden units (4096 in feedforward layers). 
In addition to the dropout of 0.3 applied to attention and ReLU activation, outputs from the embedding layer and the last layer had a dropout of 0.2. 
We used the adaptive input~\cite{baevski2018adaptive} and the adaptive softmax~\cite{grave2017efficient} for reducing the number of parameters within word embeddings.
The models are trained for 300k updates with a block size of 256, and gradients are clipped at 0.1.
The other hyperparameters are the same as the small Enwik8 experiments.

\end{document}